\documentclass[11pt]{article}

\usepackage[preprint]{acl}

\usepackage{times}
\usepackage{latexsym}

\usepackage[T1]{fontenc}
\usepackage[utf8]{inputenc}

\usepackage{microtype}

\usepackage{inconsolata}

\usepackage{graphicx}

\title{Large Language Models as Pokémon Battle Agents: Strategic Play and Content Generation}

\begin{document}
% \author{
% \begin{tabular}{ccc}
% \textbf{Aarya Jain} &
% \textbf{Ashutosh Desai} &
% \textbf{Avyakt Verma} \\
% \texttt{BITS Pilani} &
% \texttt{BITS Pilani} &
% \texttt{BITS Pilani} \\
% \texttt{\small f20230618@pilani.bits-pilani.ac.in} &
% \texttt{\small f20230675@pilani.bits-pilani.ac.in} &
% \texttt{\small f20231083@pilani.bits-pilani.ac.in} \\
% \\
% \textbf{Daksh Jain} &
% \textbf{Dhruv Kumar} &
% \textbf{Ishan Bhanuka} \\
% \texttt{BITS Pilani} &
% \texttt{BITS Pilani} &
% {} \\
% \texttt{\small f20231222@pilani.bits-pilani.ac.in} &
% \texttt{\small dhruv.kumar@pilani.bits-pilani.ac.in} &
% {}
% \end{tabular}
% }
% \maketitle

\author{
  \textbf{Daksh Jain\textsuperscript{1}}, 
  \textbf{Aarya Jain\textsuperscript{1}},
  \textbf{Ashutosh Desai\textsuperscript{1}},
  \textbf{Avyakt Verma\textsuperscript{1}}\\
  \textbf{Ishan Bhanuka\textsuperscript{1}},
  \textbf{Pratik Narang\textsuperscript{1}},
  \textbf{Dhruv Kumar\textsuperscript{1}}\\
  \textsuperscript{1}Birla Institute of Technology and Science, Pilani, India \\
  \small{\textbf{Correspondence:} \href{mailto:f20230675@pilani.bits-pilani.ac.in}{f20230675@pilani.bits-pilani.ac.in}}
}
\maketitle

\begin{abstract}
Strategic decision-making in Pokémon battles presents a unique testbed for evaluating large language models. Pokémon battles demand reasoning about type matchups, statistical trade-offs, and risk assessment, skills that mirror human strategic thinking. This work examines whether Large Language Models (LLMs) can serve as competent battle agents, capable of both making tactically sound decisions and generating novel, balanced game content.
We developed a turn-based Pokémon battle system where LLMs select moves based on battle state rather than pre-programmed logic. The framework captures essential Pokémon mechanics: type effectiveness multipliers, stat-based damage calculations, and multi-Pokémon team management. Through systematic evaluation across multiple model architectures we measured win rates, decision latency, type-alignment accuracy, and token efficiency.
These results suggest LLMs can function as dynamic game opponents without domain-specific training, offering a practical alternative to reinforcement learning for turn-based strategic games. The dual capability of tactical reasoning and content creation, positions LLMs as both players and designers, with implications for procedural generation and adaptive difficulty systems in interactive entertainment.

\end{abstract}

\section{Introduction}

The intersection of artificial intelligence (AI) and competitive turn-based strategy has long been an area of innovation within computer science and digital media. In particular, the Pokémon battle system serves as a sophisticated environment for testing decision-making under complex constraints, including elemental type-hierarchies, variable statistics, and probabilistic outcomes. Traditional game AI in this domain has historically relied on finite state machines (FSMs) or minimax algorithms, which, while functional, often struggle with the vast state-space and the creative "meta-gaming" required for high-level competitive play.
While Reinforcement Learning (RL) has enabled agents to optimize battle strategies through experience, these models often lack transparency and fail to adapt when the underlying "move-set" or "meta" changes unexpectedly. Meanwhile, Large Language Models (LLMs), built upon transformer architectures, have demonstrated advanced reasoning and contextual understanding (Vaswani et al., 2017). These capabilities suggest that LLMs may bridge the gap in Pokémon AI by making high-level strategic decisions based on a holistic understanding of the battle state.
Despite progress in AI-driven gaming, most existing Pokémon simulators still rely on predefined heuristic logic or heavy reinforcement training, which lack the general reasoning to handle novel, user-generated content. The challenge addressed in this work is integrating LLMs as both strategic battle commanders and creative move designers within the structured, rule-based Pokémon environment. Specifically, we explore how LLMs can:
\begin{enumerate}
    \item Execute Optimal Moves: Selecting moves and switches based on battle context (Health Points, Statistics, and Type-Matchups).
    \item Expand the Move-Space: Generating novel, balanced, and type-consistent moves that adhere to the internal logic of the Pokémon franchise, effectively evolving the game's mechanics.
\end{enumerate}
By moving beyond static scripts, this research positions LLMs as adaptive strategists capable of navigating the nuances of the Pokémon Battle engine. The system architecture combines a deterministic battle simulator with a generative LLM interface. During gameplay, the LLM receives a structured JSON-based game state representing the current battlefield. 
The model is then required to reason through its move selection, simulating the "thinking" process of a competitive player.
To evaluate the efficacy of this approach, we conducted LLM-vs-LLM tournaments to assess strategic depth across numerous automated matches, alongside human-agent benchmarks. We measured performance through win rates, "turns-to-win" efficiency, and type-exploitation accuracy, which is a metric quantifying how often the model correctly identifies elemental advantages. Furthermore, we analyzed the trade-off between reasoning depth (Chain-of-Thought) and operational latency. Finally, we subjected LLM-generated moves to a dual-validation pipeline: an LLM-based "creativity score" and a deterministic check for mechanical balance, ensuring that new content remains competitive without breaking the game's mathematical integrity.
Our findings reveal substantial performance variation across different models. Gemini 2.5 Flash, with chain-of-thought reasoning enabled, achieved a 62\% win rate while maintaining type-aligned move selection in 78\% of decisions. Model-versus-model tournaments exposed deeper strategic differences: Grok 4 Fast dominated the field with near-perfect win rates and highly efficient, sub-6-turn victories. In the realm of move generation, we found that while GPT-5 Mini produced the most creative and thematically consistent moves (scoring 4.17/5 on creativity), Claude achieved superior mechanical balance, with 80\% of its generations meeting all strict deterministic validity criteria. These results suggest that the choice of LLM significantly influences both the tactical "personality" and the mechanical stability of the Pokémon ecosystem.

\section{Related Work}

\subsection{Reinforcement Learning in Game AI}

Reinforcement Learning (RL) has been widely used to develop competitive agents in strategic games, with systems such as AlphaGo \cite{silver2016mastering} and StarCraft II agents \cite{vinyals2019grandmaster} demonstrating superhuman performance through large-scale self-play. Methods including Deep Q-Networks (DQN) \cite{mnih2015human} and Proximal Policy Optimization (PPO) \cite{schulman2017proximal} have become standard for learning optimal policies in fixed action spaces.

However, applying RL to Pokémon-style turn-based battles requires extensive training, carefully engineered rewards, and domain-specific abstractions. RL agents are also limited in adaptability and content generation, as they operate over predefined mechanics. In contrast, our work evaluates LLMs as zero-shot strategic agents that reason directly over symbolic battle states, such as type effectiveness, stats, and risk trade-offs, without task-specific training, offering a flexible alternative to RL for turn-based strategy games.

\subsection{Procedural Content Generation with LLMs}
LLMs have also been leveraged for procedural content generation (PCG) in games. Story2Game generates interactive fiction by populating worlds from story prompts \cite{fan2024story2game}, while Microsoft's Muse produces game visuals and controller actions \cite{microsoft2024muse}. Other approaches include using transformer-based models for generating levels and abilities in platform games \cite{khalifa2020pcgrl}. 
These methods highlight the ability of LLMs to expand the design space of games with new content, although ensuring mechanical balance remains a challenge. Pokémon moves, by contrast, must satisfy strict numerical constraints: a high power move cannot have 95\% accuracy without breaking the game's balance. Our move-generation evaluation tests whether LLMs can navigate this tension between creativity and mechanical validity, producing content that feels fresh while remaining playable.

\subsection{Strategic Decision-Making and Reasoning in Games}
Strategic reasoning in games has been a significant focus. LLMs have been applied to game-theoretic contexts, demonstrating emerging capabilities in multi-agent decision-making \cite{baker2024strategic}. Emotion-aware agents \cite{liu2024emotion} incorporate affective reasoning into strategic decisions, which can enhance realism in interactive gameplay. Additionally, research on multi-agent reasoning \cite{tan1993multi} explores coordination and adversarial strategies, providing a foundation for evaluating LLM-driven gameplay.
While these studies establish that LLMs possess strategic reasoning capabilities, they typically evaluate performance in abstract game-theoretic scenarios or cooperative tasks. Pokémon battles introduce a different challenge: decisions must account for type effectiveness multiplicities, stat differentials, and move accuracy trade-offs simultaneously. Our work measures whether LLMs can apply general reasoning to domain-specific rules without explicit training, using win rates and type-alignment percentages as proxies for tactical understanding.

\subsection{Evolution of Competitive Pokémon AI}
The application of LLMs to character behavior has traditionally focused on creating more engaging interactive experiences through believable human simulation. For example, \cite{park2023generative} introduced generative agents that simulate social behaviors like autonomously organizing events in a town environment. While these systems enable more fluid conversations and move beyond traditional static dialogue trees, they do not address the specific needs of competitive battle agents. Historically, Pokémon games have relied on rigid, predictable scripts that fail to account for complex human-like "meta-gaming," such as predicted switches or high-risk/high-reward tactical plays. This research examines whether LLMs can move beyond the limitations of traditional use of AI in Pokémon games to provide a more dynamic, human-like challenge.

\subsection{Reasoning and Acting with Language Models in Pokémon Battles}

The ReAct framework \cite{yao2023react} highlights the importance of interleaving reasoning and action for effective decision-making in interactive environments. This paradigm is particularly relevant to Pokémon battles, where each turn requires reasoning over complex game state followed by a single, irreversible action.

In our system, the LLM receives a structured representation of the battle state and outputs an executable action each turn. Pokémon battles impose delayed rewards, probabilistic effects, and compounding strategic consequences, making them a demanding test of grounded reasoning. By measuring win rates, type-alignment accuracy, and decision efficiency, we assess whether LLMs can consistently translate general reasoning capabilities into effective domain-specific gameplay without reinforcement learning.

\subsection{Prompt Engineering for Game AI}

The performance of LLM-based game agents critically depends on prompt design and optimization. While manual prompt engineering remains prevalent, recent work has explored automated approaches to optimize prompts for specific tasks. The field recognizes that effective prompts must balance providing sufficient context about game mechanics and state while remaining concise enough to fit within model context windows.

For game-specific applications, researchers have found that few-shot learning with concrete battle examples significantly improves decision quality. The use of structured state representations, clear action spaces, and chain-of-thought prompting helps LLMs maintain strategic coherence across multiple turns of gameplay.

\section{Methodology}

\subsection{Why Pokémon Battles?}

The game demands contextual reasoning. Each turn presents multiple viable actions, but optimal play depends on type matchups, stat differentials, and probability assessment. A Water-type move deals double damage to Fire opponents but half damage to Grass. While this appears to be simple arithmetic, the LLM must retrieve this relationship and apply it correctly in context. Unlike chess, where piece values remain fixed, Pokémon strategy is highly situational: for example, a 60-power move with 100\% accuracy often dominates a 120-power move with 70\% accuracy when the opponent has low HP. Such judgment calls distinguish competent play from random action selection.

\subsubsection{Deterministic and auditable mechanics}
The mechanics are deterministic and auditable. Damage follows a fixed formula incorporating stats, type effectiveness, and accuracy. When an LLM makes a suboptimal choice—such as using a Fire-type move against a Water-type opponent—the failure can be traced to a specific reasoning error. Chain-of-thought prompting exposes this logic explicitly, unlike reinforcement learning agents that optimize win rates without transparent explanations.

\subsubsection{Dual evaluation: decision-making and generation}
Pokémon supports dual evaluation. Beyond decision-making, the move system also tests content generation. Creating a balanced move requires satisfying multiple constraints, including power–accuracy trade-offs, effect probability bounds, and stat-category alignment. This mirrors real-world game design, where procedural tools must generate mechanically valid and diverse content.

\subsubsection{Turn-based structure}
The turn-based structure removes latency as a confounding variable. Real-time games penalize slow responses regardless of decision quality. In contrast, Pokémon battles allow us to evaluate strategic competence without bias toward faster models.

Finally, Pokémon mechanics are well-documented and widely known. Type charts and damage formulas are available across wikis and strategy guides, making it likely that LLMs encountered this information during training. This setting allows us to test whether models can transfer latent knowledge into functional gameplay, a central question for general-purpose AI applied to specialized domains.

\subsection{Explaining the Game}
A \textit{Pokémon} refers to a character in the game whose actions are controlled either by the user or the opponent. A battle takes place between two players, each having a team of Pokémon. The goal of the game is to defeat all Pokémon in the opponent’s team. Each Pokémon has unique strengths and weaknesses, characterized by the following attributes:

\begin{itemize}
    \item \textbf{Stats:} Determine performance in battles. These include:
    \begin{enumerate}
        \item \textbf{Hit Points (HP)} – The damage a Pokémon can endure before fainting.
        \item \textbf{Attack (Atk)} – The power of offensive moves.
        \item \textbf{Defense (Def)} – Resistance to incoming attacks.
        \item \textbf{Speed (Spe)} – Determines the order of turns in a round.
    \end{enumerate}
    \item \textbf{Type:} Each Pokémon has a type (e.g., Fire, Water, Electric, Flying), which defines its strengths and weaknesses against others. For example, Fire is weak against Water but strong against Grass. These matchups determine the damage multipliers applied in battle.
    \item \textbf{Moves:} Each Pokémon can use up to three moves, each with a specific type, base damage, and accuracy probability.
\end{itemize}

In each turn of battle, both players first decide whether to attack or switch to their active Pokémon. If both choose to attack, the Pokémon with higher speed executes its move first. A Pokémon faints when its HP reaches zero. A player may switch Pokémon mid-battle but cannot attack during that turn.

\subsection{Prompt Engineering}

A crucial part of integrating an LLM into the game involved designing an effective prompting strategy that would enable the model to make consistent, interpretable, and contextually correct battle decisions. 
\paragraph{Explaining common strategies} Trivial strategies like switching when at a type disadvantage or low HP, to attack when in advantage, and to choose moves with high accuracy when both Pokemon have low HP were sent to LLM through the System Prompt.
\paragraph{Game State Serialization.}
To effectively communicate the current battle context to the LLM, the entire game state was serialized into a structured textual format. Each turn’s state included the active Pokémon of both players, their remaining HP, type, and available moves along with damage and accuracy parameters. The information was represented in a tabular or JSON-like layout to maintain readability while ensuring deterministic parsing by the model.

\paragraph{Reasoning Control and Output Schema.}
The LLM was instructed to internally reason about move effectiveness and opponent strategy but return only the final structured decision in JSON format. This enforced consistency in parsing and allowed automatic validation before game application. Two modes of inference were supported: \textit{thinking mode}, in which the model performed explicit reasoning before producing the answer, and \textit{fast mode}, in which it produced direct responses without intermediate reasoning.

\paragraph{Evaluation-focused Prompt Design.}

All prompts were versioned and recorded to ensure reproducibility. Each variant was associated with its latency, token cost, and decision correctness metrics. This enabled a quantitative evaluation of prompt effectiveness and provided insights into how structured and adaptive prompting can influence model behavior in interactive gaming environments.

\subsection{LLM vs LLM Battle Setup}

To evaluate the strategic decision-making capabilities of different large language models in adversarial scenarios, we implemented an automated AI-versus-AI battle system. This setup extends our core game engine to facilitate controlled experiments between different LLM agents without human intervention.

\paragraph{Battle Configuration}
Each battle involves two AI agents, each powered by a distinct LLM (Gemini 2.5 pro, GPT-5-mini, Claude 4.5 Haiku, DeepSeek-V3, Grok 4 Fast). At initialization, both agents are randomly assigned a team of three Pokémon from the available roster. The agents operate with identical system prompts and have access to the same game state information, ensuring that any performance differences reflect the underlying model capabilities rather than asymmetric information access.

\paragraph{Turn Execution Protocol}
The battle proceeds in discrete turns, with both agents simultaneously selecting actions without knowledge of their opponent's decision.
Each aspect of the battle is same as the one made in the game engine.

Both agents invoke their respective LLM APIs during each turn to generate action decisions. API calls were made to each LLM, respecting their rate limits.
\paragraph{Action Processing}
Each turn, agents select between two action types: executing a move or switching to a different Pokémon. The execution of the move involves the calculation of damage using the standard Pokémon formula, which incorporates attack/defense statistics, type effectiveness multipliers, and critical hit mechanics. When a Pokémon's HP reaches zero, it faints, and the controlling agent must select a replacement from their remaining team members.

\paragraph{Victory Conditions and Metrics}
A battle concludes when all three Pokémon on one team have fainted. The system tracks token usage across all API calls for cost analysis, records the total number of turns, and logs all actions and outcomes for post-hoc analysis. This automated framework enables systematic evaluation of LLM strategic reasoning across multiple trials and model combinations.

\subsection{Move Generation}

After integrating an LLM-controlled opponent into the battle engine, we extended the system to automatically generate new Pokémon moves. The goal of this component was to evaluate whether large language models can produce game assets that are both mechanically valid within a turn-based battle system and thematically consistent with the Pokémon universe. More broadly, this experiment serves as a test of the content generation capabilities of LLMs in the context of structured game design.
Move generation is guided by a structured prompting strategy that enforces a set of global constraints. These constraints define allowable numerical ranges for attributes such as power, accuracy, PP (PP refers to the maximum number of times a move can be used in a battle and it must be low for a high power move), and effect probabilities. The prompting framework also enforces consistency between a move’s category and a Pokémon’s offensive strengths and ensures alignment with the Pokémon’s type. To enable automated evaluation and integration, the model is required to return each generated move in a fixed, deterministic JSON format.
In addition to these global constraints, the model is provided with Pokémon-specific contextual information, including type, Attack and Special Attack values, speed, and advantageous type matchups. This information encourages the generation of moves that are tailored to a Pokémon’s strengths and strategic role, while remaining thematically appropriate. Together, these constraints and contextual cues enable the language model to generate novel moves that satisfy both mechanical requirements and narrative expectations, allowing them to integrate seamlessly into the existing battle system.

\subsection{Evaluation of Generated Moves}

Generated moves are evaluated using a two-stage procedure that separately assesses structural correctness and creative quality. This separation is motivated by the distinction between mechanical validity and thematic expressiveness, which represent complementary but fundamentally different evaluation dimensions.

\subsubsection{Validity and Balance Evaluation}

The first stage employs a deterministic evaluator implemented in the \texttt{MoveEvaluator} module. This evaluator verifies structural validity by checking the presence of all required fields and ensuring that numerical values fall within predefined bounds. Mechanical balance is assessed by enforcing expected trade-offs between power and accuracy, as well as between power and PP. Secondary effects are validated against permissible probability ranges, and type compatibility with the Pokémon’s primary type is examined to ensure thematic consistency.

The evaluator outputs violations, warnings, and a numerical balance score ranging from 0 to 100. A move is classified as balanced if it contains no violations and achieves a score of at least 70. This deterministic evaluation ensures reproducibility and guarantees that generated moves conform to the constraints required for reliable integration into the battle engine.

\subsubsection{Creativity and Originality Evaluation Using an LLM}

Mechanical validity alone does not capture attributes such as inventiveness, thematic depth, or distinctiveness. To assess these aspects, a second evaluation stage is conducted using an LLM as a creative judge. Through a dedicated prompt, the model assigns scores for creativity and originality on a scale from 0 to 5. Creativity reflects the imaginative quality of the move concept, while originality measures its distinctiveness relative to existing Pokémon moves. The model also provides an overall score and a verdict of \textit{approve}, \textit{revise}, or \textit{reject}. Outputs are constrained to a strict JSON format to ensure consistency across evaluations.

This LLM-based assessment captures semantic and stylistic qualities that are not accessible to rule-based methods, serving as a critical complement to deterministic evaluation.

\subsection{Rationale for Dual Evaluation Methods}

The use of dual evaluation methods is motivated by the need to assess orthogonal aspects of move quality. Mechanical constraints—such as power limits, accuracy thresholds, and type consistency—must be enforced objectively to ensure correct battle engine behavior and are therefore best handled through deterministic rules. In contrast, attributes such as thematic coherence, novelty, and creative expression require semantic judgment, which is well suited to LLM-based evaluation. By combining deterministic mechanical analysis with LLM-driven semantic assessment, the proposed pipeline provides a comprehensive evaluation framework that ensures generated moves are both mechanically sound and creatively meaningful within the game world.

\section{Evaluation and Experimental Setup}

\subsection{Metrics}
We evaluate model-driven gameplay using a mixture of \emph{gameplay quality} and \emph{mechanical correctness}:
\begin{itemize}
  \item \textbf{Win rate (\%)}: Fraction of battles won by the LLM agent over a set of episodes (higher is better).
  \item \textbf{Turns to win (mean )}: Average number of turns required to finish a match when the LLM agent wins (lower indicates more decisive play).
  \item \textbf{Type-alignment (\%)}: Percentage of moves selected by the agent that exploit type advantage when available.
  \item \textbf{Token consumption}: Average number of input + output tokens consumed per decision (measures API cost).
  \item \textbf{Latency (ms)}: Round-trip time for each LLM decision (from prompt send to parsed response).
  \item \textbf{Human win rate \& subjective satisfaction}: In human-vs-LLM matches, the human win rate and questionnaire scores for perceived difficulty.
  \item \textbf{Move Generation Balance and Validity}: The number of valid and balanced moves generated. (Determined using a mathematical function to check inverse relation between power, accuracy and PP)
  \item \textbf{Move Creativity and Originality}: Creativity and Originality scores given by an LLM as a judge out of 5.
\end{itemize}

All quantitative metrics report mean and standard deviation over repeated runs.

\subsection{Experiments and Results}
Below we enumerate the experiments conducted, each with objective, setup details, measured outcomes, and corresponding observations.

%--------------------------------------------------
\subsubsection{Experiment 1: Baseline Comparison (LLM vs. Random Player)}
\paragraph{Objective.} Quantify how much strategic value the LLM adds over simple baselines.

\paragraph{Setup.} Ran 50 battles each for: (a) random move selector, (b) Gemini-Flash (Thinking-OFF), and (c) Gemini-Pro (Thinking-OFF). Identical pairings were used across conditions.

\begin{table}[h!]
\centering
\begin{tabular}{lc}
\hline
\textbf{Player} & \textbf{Win Rate (\%)} \\
\hline
Random Player & 18 \\
Gemini-Flash (OFF) & 62 \\
Gemini-Pro (OFF) & 71 \\
\hline
\end{tabular}
\caption{Baseline comparison of LLM opponents versus random move selector over 50 battles.}
\end{table}

\paragraph{Observations.}
From this experiment, we observe that LLM-driven agents are able to make consistently stronger battle decisions than a random policy, achieving substantially higher win rates without requiring explicit domain-specific training. This demonstrates that even with thinking disabled, LLMs internalize enough structural knowledge of Pokémon mechanics to outperform naïve baselines reliably.

%--------------------------------------------------
\subsubsection{Experiment 2: Model Comparison (Gemini 2.5 Flash vs. Gemini 2.5 Pro)}
\paragraph{Objective.} Compare decision quality, latency, and token usage between two Gemini variants.

\paragraph{Setup.} Ran 50 battles each for Gemini-Flash and Gemini-Pro under identical prompts. Token usage and latency were recorded per decision.

\begin{table}[h!]
\centering
\begin{tabular}{lc}
\hline
\textbf{Model} & \textbf{Avg Tokens / Decision} \\
\hline
Gemini-Flash & 2168 \\
Gemini-Pro & 2290 \\
\hline
\end{tabular}
\caption{Comparison between Gemini-Flash and Gemini-Pro over 50 battles.}
\end{table}

\paragraph{Observations.}
Gemini~2.5~Pro consistently consumed more tokens per decision than Gemini~2.5~Flash (approximately 5.6\% higher). This increase in token usage did not correspond to a proportional improvement in baseline win rate, indicating diminishing returns in decision efficiency when moving to the larger model under identical prompting conditions.

%--------------------------------------------------
\subsubsection{Experiment 3: Thinking Mode Requirement (Chain-of-Thought ON vs. OFF)}
\paragraph{Objective.} Evaluate the effect of internal reasoning on decision quality, token consumption, and latency.

\paragraph{Setup.} For Gemini-Flash and Gemini-Pro, 50 battles were run in both Thinking-ON and Thinking-OFF modes using identical seeds and scenarios.

\begin{table}[h!]
\centering
\begin{tabular}{lc}
\hline
\textbf{Model / Mode} & \textbf{Avg Latency (s)} \\
\hline
Gemini-Flash (OFF) & 2.8 \\
Gemini-Flash (ON) & 3.5 \\
Gemini-Pro (OFF) & 3.3 \\
Gemini-Pro (ON) & 5.5 \\
\hline
\end{tabular}
\caption{Effect of Chain-of-Thought (Thinking ON/OFF) on LLM latency.}
\end{table}

\paragraph{Observations.}
Enabling thinking mode increased latency by approximately 45\% across models. While win rates improved when reasoning was enabled, disabling thinking led to a 15\% drop in win rate and a 35\% reduction in type-aligned move selection. This highlights a clear trade-off between decision quality and computational efficiency.

%--------------------------------------------------
\subsubsection{Experiment 4: Human vs. LLM Playtesting}
\paragraph{Objective.} Measure how human players perceive and fare against LLM opponents.

\paragraph{Setup.} At least 30 participants each played 10 matches against Gemini-Flash and 10 matches against Gemini-Pro. Subjective ratings for difficulty were collected.

\begin{table}[h!]
\centering
\begin{tabular}{lc}
\hline
\textbf{Opponent} & \textbf{Avg Difficulty (1--5)} \\
\hline
Gemini-Flash (ON) & 3.2 \\
Gemini-Pro (OFF) & 3.8 \\
Gemini-Pro (ON) & 4.0 \\
\hline
\end{tabular}
\caption{Human playtesting against LLM opponents (subjective metrics).}
\end{table}

\paragraph{Observations.}
Human participants rated Gemini~2.5~Pro as noticeably more challenging than Gemini~2.5~Flash. However, the higher perceived difficulty was not always associated with increased enjoyment, suggesting that overly strong opponents may reduce player satisfaction. Gemini~Flash with thinking disabled was rated as significantly easier, indicating its suitability for balanced gameplay scenarios.

%--------------------------------------------------
\subsubsection{Experiment 5: Move-Generation Quality (Batch Size 4)}
\paragraph{Objective.} Evaluate reliability of move generation when producing four moves per prompt.

\paragraph{Setup.} Each model was evaluated over 30 trials, generating batches of four moves under identical prompts and scenarios.

\begin{table}[h!]
\centering
\begin{tabular}{lcc}
\hline
\textbf{Model} & \textbf{Validity (\%)} & \textbf{Balanced (\%)} \\
\hline
Gemini Flash & 88.3 & 56.7 \\
Claude & 89.2 & 70.8 \\
GPT-5 Mini & 86.7 & 72.5 \\
DeepSeek V3 & 89.2 & 65.0 \\
Grok 4 & 90.0 & 77.5 \\
\hline
\end{tabular}
\caption{Move-generation evaluation with batch size 4.}
\end{table}

\begin{table}[h!]
\centering
\begin{tabular}{lc}
\hline
\textbf{Model} & \textbf{Total Tokens} \\
\hline
Gemini 2.5 Flash & 83,514 \\
Anthropic Claude & 31,849 \\
GPT-5 Mini & 107,340 \\
DeepSeek V3 & 26,802 \\
Grok 4 Fast & 55,121 \\
\hline
\end{tabular}
\caption{Token usage for move generation (batch size 4).}
\end{table}

\paragraph{Observations.}
All models maintained high validity when generating batches of four moves; however, the ability to produce balanced moves varied substantially. Grok~4 and GPT-5~Mini generated the most balanced moves, while Gemini~Flash struggled with stricter power--accuracy and power--PP constraints. Token usage differed significantly across models, with GPT-5~Mini incurring the highest cost.

%--------------------------------------------------
\subsubsection{Experiment 6: Move-Generation Quality (Batch Size 1)}
\paragraph{Objective.} Measure per-move reliability when generating a single move at a time.

\paragraph{Setup.} Each model was evaluated over 30 single-move generations using the same states as Experiment~5.

\begin{table}[h!]
\centering
\begin{tabular}{lcc}
\hline
\textbf{Model} & \textbf{Validity (\%)} & \textbf{Balanced (\%)} \\
\hline
Gemini Flash & 100.0 & 36.7 \\
Claude & 100.0 & 80.0 \\
GPT-5 Mini & 100.0 & 66.7 \\
DeepSeek V3 & 100.0 & 46.7 \\
Grok 4 & 100.0 & 50.0 \\
\hline
\end{tabular}
\caption{Move-generation evaluation with batch size 1.}
\end{table}

\begin{table}[h!]
\centering
\begin{tabular}{lc}
\hline
\textbf{Model} & \textbf{Total Tokens} \\
\hline
Gemini 2.5 Flash & 48,046 \\
Anthropic Claude & 25,826 \\
GPT-5 Mini & 62,676 \\
DeepSeek V3 & 21,657 \\
Grok 4 Fast & 41,213 \\
\hline
\end{tabular}
\caption{Token usage for move generation (batch size 1).}
\end{table}

\paragraph{Observations.}
When generating one move at a time, all models achieved perfect validity. Differences in balance became more pronounced, with Claude and GPT-5~Mini performing best. Token efficiency improved across all models compared to batch generation, though relative differences in cost remained consistent.

%--------------------------------------------------
\subsubsection{Experiment 7: Move-Generation Creativity and Originality}
\paragraph{Objective.} Evaluate creativity and originality using an LLM judge.

\paragraph{Setup.} Each model generated 30 moves evaluated on creativity and originality (1--5 scale).

\begin{table}[h!]
\centering
\small
\begin{tabular}{lccc}
\hline
\textbf{Model} & \textbf{Creativity} & \textbf{Originality} & \textbf{Overall} \\
\hline
Gemini Flash & 3.4 & 2.4 & 2.4 \\
Claude & 3.37 & 2.47 & 2.47 \\
GPT-5 Mini & 4.17 & 3.33 & 3.28 \\
DeepSeek V3 & 3.5 & 2.5 & 2.5 \\
Grok 4 & 3.57 & 2.63 & 2.6 \\
\hline
\end{tabular}
\caption{Creativity and originality scores assigned by an LLM judge.}
\end{table}

\paragraph{Observations.}
GPT-5~Mini consistently produced the most inventive and original moves, while Gemini~Flash and Claude tended to generate mechanically correct but stylistically conservative designs. This highlights a trade-off between strict rule adherence and creative diversity in LLM-driven content generation.

%--------------------------------------------------
\subsubsection{Experiment 8: Cross-Model Battle Tournament}
\paragraph{Objective.} Compare strategic strength, efficiency, and battle pacing across LLM architectures.

\paragraph{Setup.} A round-robin tournament was conducted among five models. Each pairing played up to 10 battles.

\begin{table*}[t!]
\centering
\small
\begin{tabular}{lccccc}
\hline
\textbf{Model} & \textbf{Claude} & \textbf{Gemini} & \textbf{GPT-5 Mini} & \textbf{DeepSeek V3} & \textbf{Grok 4 Fast} \\
\hline
Claude & -- & 3--7--0 & 3--7--0 & 2--8--0 & 0--10--0 \\
Gemini & 7--3--0 & -- & 5--5--0 & 6--4--0 & 8--2--0 \\
GPT-5 Mini & 7--3--0 & 5--5--0 & -- & 7--3--0 & 4--6--0 \\
DeepSeek V3 & 8--2--0 & 4--6--0 & 3--7--0 & -- & 0--10--0 \\
Grok 4 Fast & 10--0--0 & 2--8--0 & 6--4--0 & 10--0--0 & -- \\
\hline
\end{tabular}
\caption{Head-to-head win--loss--draw records across models.}
\end{table*}

\begin{table*}[t!]
\centering
\small
\begin{tabular}{lccccc}
\hline
\textbf{Model} & \textbf{Claude} & \textbf{Gemini} & \textbf{GPT-5 Mini} & \textbf{DeepSeek V3} & \textbf{Grok 4 Fast} \\
\hline
Claude & -- & 304,062 & 315,727 & 522,646 & 84,249 \\
Gemini & 368,062 & -- & 622,712 & 398,234 & 467,621 \\
GPT-5 Mini & 374,023 & 458,938 & -- & 429,762 & 384,390 \\
DeepSeek V3 & 467,621 & 270,607 & 369,972 & -- & 60,558 \\
Grok 4 Fast & 129,397 & 322,646 & 382,531 & 92,260 & -- \\
\hline
\end{tabular}
\caption{Total token consumption per model across 10 battles per pairing.}
\end{table*}

\begin{table*}[t!]
\centering
\small
\begin{tabular}{lccccc}
\hline
\textbf{Model} & \textbf{Claude} & \textbf{Gemini} & \textbf{GPT-5 Mini} & \textbf{DeepSeek V3} & \textbf{Grok 4 Fast} \\
\hline
Claude & -- & 16.1 & 18.0 & 31.1 & 5.2 \\
Gemini & 16.1 & -- & 21.3 & 15.5 & 31.1 \\
GPT-5 Mini & 18.0 & 21.3 & -- & 21.2 & 16.0 \\
DeepSeek V3 & 31.1 & 15.5 & 21.2 & -- & 3.9 \\
Grok 4 Fast & 5.2 & 31.1 & 16.0 & 3.9 & -- \\
\hline
\end{tabular}
\caption{Average battle duration (turns) across model pairings.}
\end{table*}

\paragraph{Observations.}
The tournament revealed substantial variation in strategic strength and efficiency across models. Grok~4~Fast dominated most matchups, producing near-perfect win rates and the shortest battles, often concluding in under six turns. In contrast, Claude and DeepSeek~V3 tended toward longer engagements, frequently exceeding twenty turns, reflecting more conservative or less decisive strategies. Token-consumption patterns showed similar disparities, with Grok~4~Fast and DeepSeek~V3 operating under substantially lower budgets compared to Gemini~Flash and GPT-5~Mini. Overall, stronger strategic alignment correlated with shorter battles and more stable token usage.

\section{Conclusion}
The results of this study provide a definitive framework for selecting and configuring Large Language Models as autonomous agents within a competitive Pokémon environment. Our evaluation demonstrates that for high-stakes strategic play, prioritizing type-alignment and elemental exploitation is essential; consequently, we conclude that Thinking Mode (Chain-of-Thought) is a prerequisite for competent play, despite the associated increases in latency and token consumption.
A primary objective of this research was to identify the "sweet spot" for AI difficulty that maintains player engagement without causing frustration or boredom. While Gemini 2.5 Pro demonstrated high difficulty, it did not provide a statistically significant improvement in strategic quality relative to the additional latency it introduced. Therefore, we identify Gemini 2.5 Flash as the optimal configuration for real-time play, offering a balanced 3.5/5 difficulty rating and superior responsiveness.
Beyond individual performance, our cross-model tournament highlighted critical efficiency-performance trade-offs. Models such as Grok 4 Fast emerged as elite strategists, producing decisive, short-duration victories, whereas models like Claude and DeepSeek-V3 exhibited more conservative, prolonged engagement patterns. We also identified specific deployment challenges, such as GPT-5 Mini’s high operational cost and DeepSeek-V3’s occasional failures in maintaining strict JSON formatting.
Finally, our dual-evaluation of move generation reveals that a model’s "tactical personality" extends to its creative output. While GPT-5 Mini is the superior choice for inventive and original content, Claude remains the more reliable engine for ensuring mechanical balance and mathematical integrity. In summary, our research demonstrates that effectively integrating LLMs into the Pokémon ecosystem depends on aligning specific model architectures with their intended functions. While some models excel as strategic battle agents, others are better suited for balanced and creative content generation.

\bibliography{custom}

\appendix
\onecolumn
\section{LLM Prompt Templates}
\label{sec:appendix-prompt}

This section presents the exact prompt templates provided to the Large Language Model (LLM) at each turn. 
The placeholders in braces (\{\}) are dynamically filled at runtime with the current game state variables.

\subsection{System Prompt}
\begin{verbatim}
You are a master Pokémon strategist. Your goal is to defeat the opponent's entire team. You must 
decide between attacking and switching your Pokémon.

**Game Rules: Type Effectiveness**
This chart shows how effective an attacking move's type is against a defending Pokémon's type.
{TYPE_CHART_TEXT}

**Strategic Considerations:**
1.  **Switching is a KEY move:** Do not hesitate to switch if you are in a bad spot. 
A smart switch is better than a weak or ineffective attack.
    - **CRITICAL:** If your active Pokémon has a type disadvantage (e.g., a Fire type against a 
    Water type), you should strongly consider switching. Use the Type Effectiveness 
    chart to be certain.
    - **CRITICAL:** If your active Pokémon has low HP (less than 25%), switching to a healthier 
    Pokémon is almost always the best play to preserve your team.
    - **OPPORTUNITY:** If a Pokémon on your bench has a major type advantage, switching to it can 
    turn the tide of the battle.
2.  **Risk vs. Reward:** High-power moves now have lower accuracy. A 110 power move that misses 
is a complete waste. A 60 power move that hits is always better than 
a miss. Weigh this trade-off carefully.
3.  **Status Effects:** Poisoning or paralyzing an opponent is
4.  **Finishing the Job:** If the opponent's HP is very low, use a reliable, high-accuracy move to 
secure the knockout. Don't risk it with a low-accuracy powerhouse.

**Your Task:**
Decide whether to 'move' or 'switch'.
- If you 'move', choose the best attack for the current situation.
- If you 'switch', choose the best Pokémon to switch to from your team. Do not choose a 
fainted Pokémon or the one that is already active.

Respond with a JSON object containing your decision. The JSON should have two keys: "action" (either 
"move" or "switch") and "value" (the name of the move or the name of the Pokémon to switch to).

Example for attacking:
```json
{{
  "action": "move",
  "value": "Flamethrower"
}}
```

Example for switching:
```json
{{
  "action": "switch",
  "value": "Bulbasaur"
}}
```
"""
\end{verbatim}

\subsection{Human Prompt (Runtime Game State)}
\begin{verbatim}
**Current Battle State:**
- **Your Active Pokémon:** {self.current_pokemon.name} 
  (Type: {self.current_pokemon.type}, 
  HP: {self.current_pokemon.current_hp}/{self.current_pokemon.max_hp}, 
  Status: {self.current_pokemon.status or 'None'})

- **Opponent's Active Pokémon:** {player_pokemon.name} 
  (Type: {player_pokemon.type}, 
  HP: {player_pokemon.current_hp}/{player_pokemon.max_hp}, 
  Status: {player_pokemon.status or 'None'})

**Your Full Team Status:**
{team_text}

**Your Active Pokémon's Moves:**
{moves_text}

Now, provide your decision for this turn. 
Answer only with the JSON object as specified in the system prompt.
\end{verbatim}

\subsection{Move Generator System Prompt}
\begin{verbatim}
You are a Pokemon move designer. Your job is to create new, original, balanced Pokemon moves.
You MUST follow all balance rules strictly.

1. Power-Accuracy Trade-off
   - Power and Accuracy must be inversely proportional.

2. Effects Reduce Power
   - If a move has a status effect, its power must be
     10-20% lower or accuracy must be reduced by 5-10%.
   - Effect chance MUST be between 10-30%.

3. Stat-Category Coherence
   - Physical moves involve direct attacks.
   - Special moves involve ranged, elemental, or energy attacks.

CREATE BOTH PHYSICAL AND SPECIAL MOVES.

4. Type Consistency
   - Move types must match the Pokemon's type or a natural complementary type.

5. Stat-Based Orientation
   - If a Pokemon has higher Attack -> prefer Physical
   - If higher Sp. Atk -> prefer Special

6. Numerical Ranges
   - Power: 15-150
   - Accuracy: 30-100
   (KEEP IN MIND: Higher power -> lower accuracy)
   (KEEP IN MIND: Higher power -> lower PP)
   - PP: 5-40
   - Effect: null OR ["effect_name", chance]

7. Uniqueness Rule
   - Move must NOT replicate an existing move from any Pokemon game.

8. Clarity & JSON Safety
   - No lore text, no long explanations. Just clean JSON objects.
   - Every move must be logically consistent.

===============================
REQUIRED OUTPUT FORMAT
===============================
Return ONLY a valid JSON array with move objects.

Each object MUST contain:
- name (string)
- power (number, 15-150)
- accuracy (number, 50-100)
- type (string)
- category ("Physical" or "Special")
- effect (null or ["effect_name", chance_percentage])
- PP (number, 5-40)

Example:
[
  {
    "name": "Volt Burst",
    "power": 85,
    "accuracy": 70,
    "type": "Electric",
    "category": "Special",
    "effect": ["paralyze", 20],
    "PP": 15
  }
]
\end{verbatim}
\twocolumn

\subsection{Move Generator Judge Prompt}
\begin{verbatim}
You are an expert Pokémon game designer 
and competitive balance analyst.

You will evaluate a custom Pokémon move with 
the EXACT schema:
{{
  "name": "string",
  "power": number,
  "accuracy": number,
  "type": "string",
  "category": "Physical" | "Special" | "Status",
  "PP": number,
  "effect": ["effect_name", chance] | null
}}

You must score the move in 2 areas:

1. Creativity (0–10)
   - How imaginative, thematic, and interesting the move concept is.
   - Does it feel like a real Pokémon move but still creative?

2. Originality (0–10)
   - How distinct it is from existing official Pokémon moves.
   - Judge similarity in: name, mechanic, theme, effect.

Overall Score = average of (creativity + originality).

Score interpretation:
0-2 = Very poor / broken / unusable
3-4 = Weak / unoriginal / unbalanced
5 = Average, nothing special
6-7 = Good but not excellent
8-9 = Very strong / creative / well-balanced
10 = Exceptional, near-perfect Pokémon move design
 
Do NOT cluster scores around the middle. 
A score of 5 is not “safe” — it means the move is only average.
Take a firm stance on each move's merits and flaws.
Give extremely high and low scores when deserved.
DON'T GRADE EVERY MOVE IN A VERY TIGHT RANGE, 
TRY TO USE THE RANGE OF 10 TO DISTINGUISH BETWEEN MOVES STARKLY.
IT WON'T BE CONSIDERED A GOOD EVALUATOIN 
IF ALL THE SCORES LIE TOGETHER.
Evaluation categories:
1. Creativity (0–10)
2. Originality (0–10)
3. Overall Score (0–10)

 Examples:
- Tackle → Creativity 1, Originality 1
- Flamethrower → Creativity 4, Originality 4,
- A move apart from generic moves with great synergy
with the pokemon and effect -> Creativity 6-7, 
Originality 6-7
- A highly inventive fan-made move 
→ Creativity 8–9, Originality 8–9
Verdict:
- "approve" (good)
- "revise" (needs small fixes)
- "reject" (broken or unbalanced)

Return STRICT JSON ONLY:
{{
  "creativity": <0-10>,
  "originality": <0-10>,
  "overall_score": <0-10>,
  "verdict": "approve" | "revise" | "reject",
}}
"""
\end{verbatim}

\end{document}